\documentclass[sn-mathphys-num]{sn-jnl}

\usepackage{graphicx}%
\usepackage{multirow}%
\usepackage{amsmath,amssymb,amsfonts}%
\usepackage{amsthm}%
\usepackage{natbib}
\usepackage{mathrsfs}%
\usepackage[title]{appendix}%
\usepackage{xcolor}%
\usepackage{textcomp}%
\usepackage{rotating}
\usepackage{manyfoot}%
\usepackage{booktabs}%
\usepackage{algorithm}%
\usepackage{algorithmicx}%
\usepackage{algpseudocode}%
\usepackage{listings}%
\usepackage{subcaption}
\usepackage{float}
\usepackage{amsmath}
\usepackage{svg}
\DeclareMathOperator*{\argmax}{argmax}
\DeclareMathOperator*{\SatAgg}{SatAgg}

\theoremstyle{thmstyleone}%
\theoremstyle{thmstyletwo}%

\theoremstyle{thmstylethree}%

\begin{document}

\title[Article Title]{OntoMedRec: Logically-Pretrained Model-Agnostic Ontology Encoders for Medication Recommendation}


\author[1]{\fnm{Weicong} \sur{Tan}}\email{weicong.tan@monash.edu}

\author*[1]{\fnm{Weiqing} \sur{Wang}}\email{teresa.wang@monash.edu}

\author[1]{\fnm{Xin} \sur{Zhou}}\email{xin.zhou@monash.edu}

\author[2]{\fnm{Wray} \sur{Buntine}}\email{wray.b@vinuni.edu.vn}

\author[3]{\fnm{Gordon} \sur{Bingham}}\email{g.bingham@alfred.org.au}

\author[4]{\fnm{Hongzhi} \sur{Yin}}\email{db.hongzhi@gmail.com}

\affil*[1]{\orgdiv{Faculty of Information Technology}, \orgname{Monash University}, \orgaddress{\state{Victoria}, \postcode{3800}, \country{Australia}}}

\affil[2]{\orgdiv{College of Engineering and Computer Science}, \orgname{Vin University}, \orgaddress{\city{Hanoi}, \country{Vietnam}}}

\affil[3]{\orgdiv{Nursing Services}, \orgname{Alfred Health}, \orgaddress{\city{Melbourne}, \state{Victoria}, \postcode{3004}, \country{Australia}}}

\affil[4]{\orgdiv{Faculty of Engineering}, \orgname{The University of Queensland}, \orgaddress{\city{Brisbane}, \state{Queensland}, \postcode{4072}, \country{Australia}}}


\abstract{Recommending medications with electronic health records (EHRs) is a challenging task for data-driven clinical decision support systems. Most existing models learnt representations for medical concepts based on EHRs and make recommendations with the learnt representations. However, most medications appear in EHR datasets for limited times (the frequency distribution of medications follows power law distribution), resulting in insufficient learning of their representations of the medications. Medical ontologies are the hierarchical classification systems for medical terms where similar terms will be in the same class on a certain level. In this paper, we propose \textbf{OntoMedRec}, the \textit{logically-pretrained} and \textit{model-agnostic} medical \textbf{Onto}logy Encoders for \textbf{Med}ication \textbf{Rec}ommendation that addresses data sparsity problem with medical ontologies.

We conduct comprehensive experiments on real-world EHR datasets to evaluate the effectiveness of OntoMedRec by integrating it into various existing downstream medication recommendation models. The result shows the integration of OntoMedRec improves the performance of various models in both the entire EHR datasets and the admissions with few-shot medications. We provide the GitHub repository for the source code\footnote{https://github.com/WaicongTam/OntoMedRec}.}




\maketitle

\section{Introduction} \label{sec:introduction}


The mass application of electronic health records (EHRs) has made data-driven clinical decision-support systems possible \cite{ai_in_healthcare}. Deep learning models designed to assist clinical practitioners in a range of tasks have emerged, with notable categories encompassing patient risk prediction, re-admission forecasting, the generation of EHR representations, and medication recommendations for prescribers. To assist medical practitioners in prescribing medications, recommending sets of medications for them accurately and efficiently has become a challenging yet crucial task. Therefore, numerous data-driven medication recommendation models have been developed, exemplified by notable solutions such as 4SDrug\cite{4sdrug}, EDGE \cite{edge}, and SafeDrug \cite{safedrug}. These models aim to predict the most suitable medication regimen based on a patient's diagnoses, medical procedures, and/or prior prescription history, as demonstrated by systems like COGNet \cite{cognet} and SARMR~\cite{sarmr}. Existing medication recommendation models fall into two categories: instance-based models and longitudinal models. Instance-based models (e.g., LEAP \cite{leap} and 4SDrug \cite{4sdrug}) recommend sets of drugs with patients' diagnoses in the current admission, whereas longitudinal models (e.g., MICRON\cite{micron}, SafeDrug \cite{safedrug} and COGNet \cite{cognet}) utilise patients' previous admissions. 

\begin{figure}[!h]
    \begin{subfigure}{0.5\textwidth}
\includegraphics[width=\linewidth]{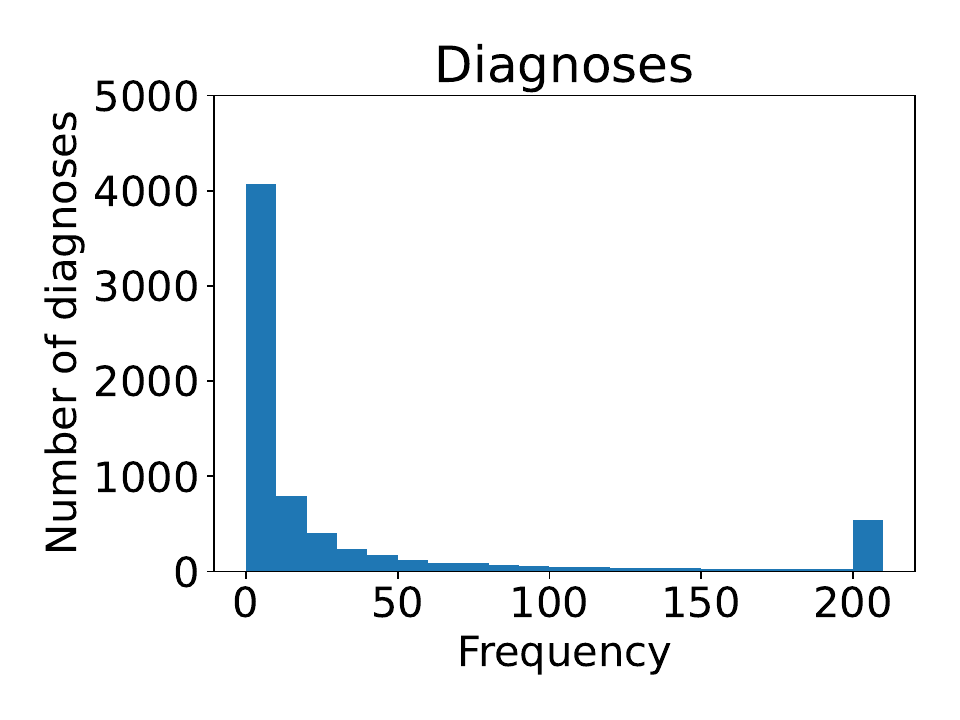} 
\label{fig:subim1}
\end{subfigure}%
\begin{subfigure}{0.5\textwidth}
\includegraphics[width=\linewidth]{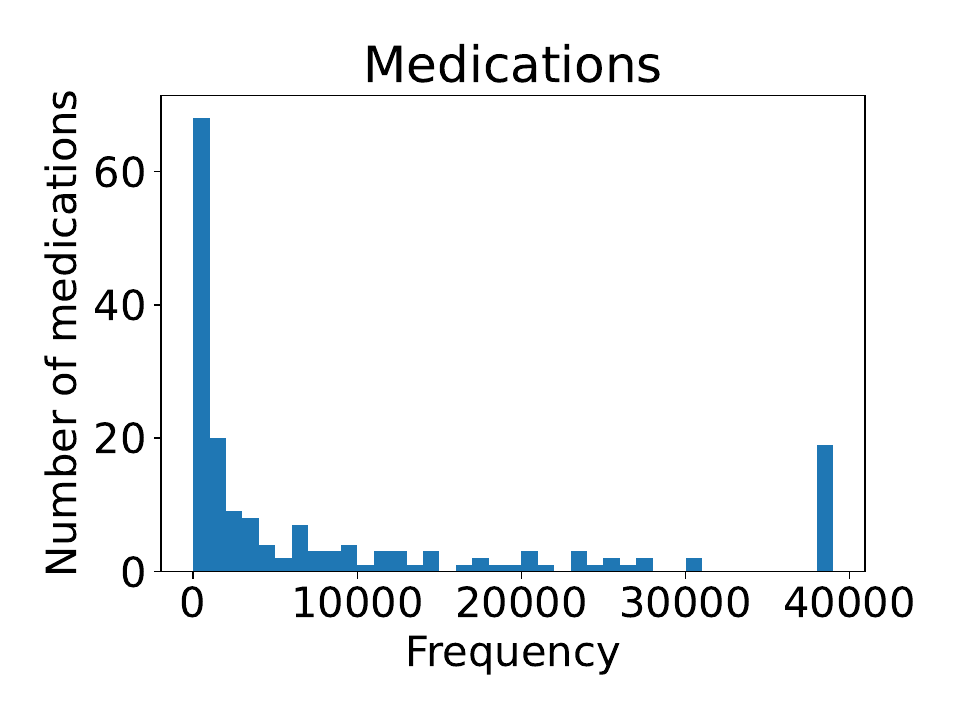}
\label{fig:subim2}
\end{subfigure}
\caption{Frequency distribution of diagnoses and medications in MIMIC-III dataset. The last bin is the cropped diagnoses/medications with a frequency higher than 200/40000.}
\label {fig:freq_dist}
\end{figure}


For both instance-based models and longitudinal medication recommendation models, we identify one challenge that has not been sufficiently addressed: \textbf{data sparsity issue (challenge 1)}. Similar to the user-interaction sparsity challenge in other recommender system models~\cite{overcoming-data-sparsity, DIRS-KG}, medication recommendation models suffer from data sparsity issues deriving from the frequency distribution of medical concepts. As demonstrated in Fig.\ref{fig:freq_dist}, the majority of diagnoses and medications only appear at limited times in the entire MIMIC-III dataset and their occurrence follows the power law distribution. This inevitably leads to insufficient learning of the indication relationships between diagnoses and medications (i.e., for what medical conditions a medication was designed) in instance-based models and their respective embeddings in longitudinal models. As proven many other recommendation tasks (e.g., \cite{RN1900} and \cite{RN1901}), utilising external knowledge bases can alleviate cold start effect. One category of the notable knowledge base for medication recommendation models is medical ontologies. Therefore, to alleviate the data sparsity issue (\textbf{challenge 1}), similar to  \cite{gbert,kampnet}, we leverage external structured knowledge (i.e., medical ontologies) \cite{gbert,kampnet} as it provides prior knowledge for the medical terms in EHRs. In EHRs, diagnoses, procedures and medications are encoded in standardised hierarchical classification systems called as medical ontologies. Each medical term is a node of the ontology and the relation between them is ``is-a'' (e.g., benproperine is a cough suppressant). 

\begin{figure}[!h]
    \centering
    \includegraphics[width=\linewidth]{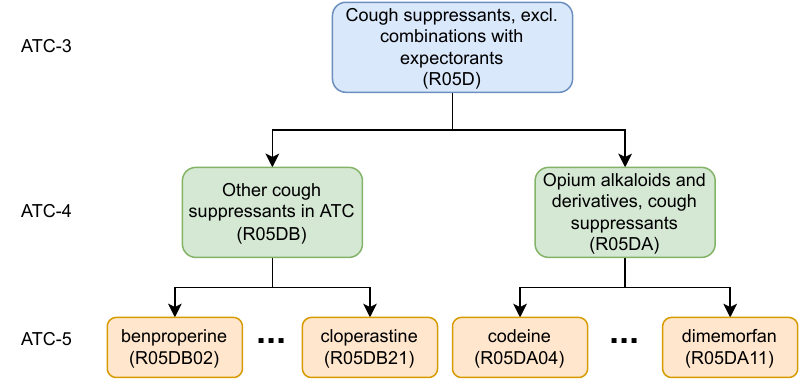}
    \caption{An excerpt of the ATC ontology. Some nodes are omitted.}
    \label{fig:atc_excerpt}
\end{figure}

Fig.\ref{fig:atc_excerpt} shows part of ATC ontology which is an ontology of medications. In this ontology, similar medications fall into the same parent node, yet there are definitive differences that distinguish them (i.e., the difference between siblings). For example, as demonstrated in Fig.\ref{fig:atc_excerpt}, medications in ``Other cough suppressant in ATC'' (R05DB) and ``Opium alkaloids and derivatives, cough suppressants'' (R05DA) fall into the same category ``Cough suppressants, excl. combinations with expectorants''(R05). However, they are intrinsically different since codeine cough suppressants (i.e., R05DA) and non-codeine cough suppressants (i.e., R05DB) have different clinical characteristics (e.g., physical dependency and drug-drug interaction). Benproperine and cloperastine have the same therapeutical classification (i.e., they are both non-codeine cough suppressants), yet they are two different chemicals. Thus, we can see from this example, that effectively modelling the parental, ancestral and sibling relationships (similarities and differences) is beneficial to the medication recommendation task.

Even though there are some works exploiting the modelling of medical ontologies in the medication recommendation task, \textbf{these existing works cannot effectively model ontology relationships to benefit medication recommendation task (challenge 2)}. Notable models integrating ontology information in medication recommendation include G-BERT \cite{gbert} and KnowAugNet \cite{kampnet}. G-BERT uses a Graph Attention Network (GAT) \cite{gat} encoder trained end-to-end along with the medication recommendation module. KnowAugNet pretrains ontology encoders with an unsupervised contrastive learning method. However, both models encode ontology with GAT and treat ontology as an undirected graph, whereas ontology is by definition a direct acyclic graph (DAG). Moreover, they cannot model some important relationships such as the difference between siblings as shown in Fig.\ref{fig:atc_excerpt}. There are also models designed for other tasks that utilised medical ontologies (e.g., GRAM \cite{gram} and KAME \cite{kame}). However, the modelling of the ontology in these methods is deeply coupled with their downstream tasks which are not medication recommendation. 

To effectively model the ontology relationships to improve the medication recommendation task (\textbf{challenge 2}), we propose a model OntoMedRec based on logic tensor networks (LTN) \cite{ltn} in this paper. As we know, LTN aims at combining symbolic rules and neural computation together. The advantage of using LTN in our task is that it allows us to easily integrate the modelling of various identified ontology relationships as symbolic rules (e.g., the parental and sibling relations in Fig.\ref{fig:atc_excerpt}) into the training process (i.e., neural computation). Recent advances in logic tensor networks (LTN) \cite{ltn} have shown its effectiveness in graph learning tasks such as ontology deduction and reasoning \cite{ltncap}. However, \textbf{we find that directly applying existing LTN technique to our task is challenging for two reasons (challenge 3)}. The first reason is that the existing LTN works are designed for different task. To adapt to medication recommendation task, we need to design new sets of predicates, axioms, constants and variables. The second reason is that directly applying existing LTN methods is memory consuming. Existing LTN studies that designed for ontology data (e.g., \cite{ltn} and \cite{ltncap}) are based on smaller ontologies (i.e., $\leq$ 100 nodes) with small representation dimensions. To model an ontology of $|\mathcal{N}|$ nodes with $n$ variables and $d$ as model dimension, the space complexity is $O(n|\mathcal{N}|d)$, which requires large amount of memory when the ontology is large (e.g., 17,737 nodes in our task). This affects the efficiency of the training process since the memories in GPUs are usually more scarce than RAMs. The high space complexity calls for an efficient sampling method for the effective training of larger node representations on larger ontologies. We devise a sampling method based on the structure of medical ontologies and our modelling method. It decreases the space complexity to $O(nbd)$ where $b << |\mathcal{N}|$ is the batch size. The contribution of this paper can be summarised as follows:
\begin{itemize}
    \item \textbf{Logically-pretrained ontology encoder: } 
    We carefully design an LTN-based encoder by devising novel predicates, axioms, constants, and variables for the self-supervised logical training on medical ontologies. \textit{The design is based on the insights of what structural information is beneficial for the medication recommendation task}. The devised axioms are naturally interpretable for humans. Moreover, for the efficient training of the model, we also designed an \textit{axiom-oriented sampling method} to enable the learning of larger node representations on large ontologies. Furthermore, to infuse the indication relationships between medications and medical diagnoses, we utilised the MEDI dataset \cite{medi} to \textit{logically align the representation space} of diagnoses and medications.

    \item \textbf{Model-agnostic ontology representation learning model for medication recommendation:} Once the encoder is well trained, its output can be loaded into various existing medication recommendation models to improve their performance as the initialisation of the embeddings of medical codes (i.e., diagnoses, procedures and/or medication embeddings). Thus, similar to other pretrained models, our encoder is \textit{``once trained and ready to use for any medication recommendation models''}. 
    
    \item \textbf{Comprehensive experiments:} 
    Comprehensive experiments have been done (with code published) to validate the effectiveness of our model in improving different existing medication recommendation methods including both \textit{instance-based methods and longitudinal methods} for both \textit{normal scenarios and few-shot scenarios}. The results show that: 1) our model is able to improve the performance of both instance-based and longitudinal downstream recommendation methods but the improvements on longitudinal methods are more obvious compared to instance-based methods; 2) our model is able to improve the performance of existing recommendation methods in both normal scenarios and few-shot scenarios but the improvement is more obvious for few-shot scenarios.
\end{itemize}

\section{Related Work} \label{sec:related}

\subsection{Instance-Based Medication Recommendation}
Instance-based models recommend a set of medications based on the current admission. LEAP \cite{leap} was an early model that predicted the prescribed medications as sequences, and it made inferences with beam search. SMR \cite{smr} recommends drugs based on knowledge graph embeddings of diagnoses and medications. More recently, 4SDrug \cite{4sdrug} was proposed. It is a set-based model trained by comparing the difference between medication sets with similar corresponding diagnoses sets. 

\subsection{Longitudinal Medication Recommendation}

Longitudinal models make use of patients' previous diagnoses and procedures records. RETAIN \cite{retain} was a representation learning model that encodes a patient's EHR into a representation, and it can be used for the medication recommendation task with extra output layers. DCw-MANN \cite{dcwmann} used all past medications to predict current medications using a LSTMs-based encoder-decoder model. GAMENet \cite{gamenet} used memory bank matrices to associate past diagnoses and procedures with medications. SafeDrug \cite{safedrug} uses a global molecule encoder and a local molecule substructure encoder to encode medications. COGNet \cite{cognet} uses Transformer-based \cite{transformer} to encode the patient's diagnoses, procedures and medication history. MICRON \cite{micron} is a model designed for predicting the change in prescriptions, it models the change of prescribed medications with residual vectors. In addition to a patient's EHR, MERITS \cite{merits} used the neural ordinary differential equation to model the irregular time series of the patient's vital signs. The model proposed by Yao et al. \cite{ontoaware} used RNN to model the path from the root node to medical concepts on medical ontologies. Other than medication recommendation, some longitudinal models use longitudinal EHR data to perform other tasks such as diagnoses prediction (e.g., KAME \cite{kame}) and representation learning (e.g., GRAM \cite{gram}). These two models also had medical ontology encoding modules, but they were trained end-to-end with downstream tasks.

\subsection{Existing Solutions to Data Sparsity Issue}
Some existing models have attempted to address the data sparsity issue. G-BERT \cite{gbert} used GAT encoders to encode diagnoses and medications. However, the pretraining data used in G-BERT is the patient records with one admission. These admission data still follow the distribution we described in Fig.\ref{fig:freq_dist}. kampnet \cite{kampnet} used unsupervised contrastive learning to pre-train encoders for medical ontologies and medication-diagnoses co-existence graph. EDGE \cite{edge} considers drugs that never appear in a certain time range in the EHR dataset as novel drugs, and uses meta-learning to alleviate the cold-start effect of those drugs. However, an interpretable and EHR-independent pretrained encoder for medical ontologies has not been proposed. Moreover, the data sparsity issue in medication recommendation has not been sufficiently addressed. 


\section{Preliminaries} \label{problem}

\subsection{Electronic Medical Record}
An electronic health record (EHR) dataset can be considered as a collection of $|\mathcal{U}|$ patients' medical records $\mathcal{U} = \{\mathcal{U}^{(n)}\}_{n=1}^{|\mathcal{U}|}$ where a patient's medical record $\mathcal{U}^{(n)}$ is constituted by their admissions $[\mathcal{V}^{(n)}_{t}]_{t=1}^{T^{(n)}}$ to the hospital. For the sake of brevity, we will omit the $(n)$ superscript in future formulae where there is no confusion. In each admission, a set of medical diagnosis codes ($\mathcal{D}_{t}$), a set of medical procedure codes ($\mathcal{P}_{t}$) and a set of prescribed medication codes ($\mathcal{M}_{t}$) will be recorded as $\mathcal{V}_{t}=\{\mathcal{D}_{t}, \mathcal{P}_{t}, \mathcal{M}_{t}\}$. Note that, in some medication recommendation models (e.g., COGNet\cite{cognet} and LEAP \cite{leap}), the set of medical diagnosis codes ($\mathcal{D}_{t}$) and the set of medical diagnosis codes ($\mathcal{D}_{t}$) are considered sequences.

It is worth noticing that the diagnosis does not only record the chief complaints (i.e., the prominent symptoms that cause this specific admission to the hospital \cite{wagner2006chief}) of the patient's admission. It also records other medical conditions of the patient. Assume that there is a diabetic patient with existing liver conditions who was admitted to the hospital due to a broken arm. Not only the bone fracture will be recorded, but the cause of the fracture (e.g., falling), the diabetes and liver conditions will be recorded as well. All the diagnosis information is codified as codes on a medical ontology that can be modelled by OntoMedRec.

\subsection{Medical Concept Ontologies}

A medical ontology $\mathcal{T}_*=\{\mathcal{N}_*, \mathcal{E}_*, \mathbf{E}_*\}$ is a hierarchical taxonomy of medical concepts in a certain domain. It is a directed acyclic graph (DAG)  where $\mathcal{N}_*$ is the set of nodes, $\mathcal{E}_*$ is the set of edges and $\mathbf{E}_*$ is the matrix of node features. An edge $e_j = \langle n_a, n_b \rangle \in \mathcal{E}_*$ represents that $n_b$ is a more specific concept deriving from $n_a$ (i.e., $n_a$ is a parent of $n_b$). Take the excerpt in Figure \ref{fig:atc_excerpt} again as an example. There is a directed edge from ``Cough suppressants, excl. combinations with expectorants (R05D)'' to ``Other cough suppressant in ATC'' (R05DB)'' since R05DB is a more specific term to classify a medication.

For OntoMedRec, we will use three non-overlapping taxonomies respectively for diagnoses, medical procedures and medications, namely $\mathcal{T}_d$, $\mathcal{T}_p$ and $\mathcal{T}_m$. Note that, $\mathcal{T}_d$, $\mathcal{T}_p$ and $\mathcal{T}_m$ are publicly available and shared by all EHR datasets by their linkage to $\mathcal{D}_t$, $\mathcal{P}_t$ and $\mathcal{M}_t$ respectively. More specifically, each medical code is a node on the corresponding medical ontology. They can be either a leaf node or a parent node. Since the medical ontology is independent of EHR datasets and all medical concepts in EHR datasets belong to the medical ontology, the pretrained representations of OntoMedRec can be integrated into any downstream recommendation models trained and tested with EHR datasets.


\subsection{Medication Recommendation}
Following the task definition in in Sec.\ref{sec:introduction} and Sec.\ref{sec:related}, the medication recommendation task can be formulated as follows:
\begin{itemize}
    \item Longitudinal models predict $\mathcal{M}_{T}$ given a patient's admission history $[\mathcal{D}_{t}, \mathcal{P}_{t}]_{t=1}^{T}$. Some of them add past medication records $[\mathcal{M}_{t}]_{t=1}^{T-1}$ (e.g., MICRON \cite{micron} and COGNet \cite{cognet}).
    \item Instance-based models predict $\mathcal{M}_{T}$ given a patient's diagnosis information in the current visit $[\mathcal{D}_{T}, \mathcal{P}_{T}]$.

\end{itemize}
 
\subsection{Logic Tensor Networks}
Logic Tensor Networks (LTNs) are the neural networks for data modelling with quantifiable and human-interpretable rules. They are based on real logic \cite{ltncap} defined on a first-order language $\mathcal{L}$. $\mathcal{L}$ is composed of \cite{ltn}:

\begin{itemize}
    \item \textbf{A set of constants.} In our case, it is the node feature matrices $\mathbf{E}_* \in \mathbb{R}^{|\mathcal{N}_*| \times d} $ where $|\mathcal{N}_*|$ is the number of nodes and $d$ is the dimension of the node representations.
    \item \textbf{A set of variables.} They are the symbols created over the subset of the constants to describe the logical relationships in the graph.
    \item \textbf{A set of predicates.} They are a set of functions $\{f_1(\cdot), f_2(\cdot), \cdots, f_n(\cdot)\}$ that take variables as inputs and calculate the satisfiability scores of a logical relationship.
    \item \textbf{A set of connectives.} They are logical operators and aggregation operators (e.g., ``and ($\land$)'' and ``not ($\neg$)''). 
\end{itemize}

There, a knowledge base can be defined as a triple $<\mathcal{K}, \mathcal{G}(\cdot|\theta), \Theta>$, where
\begin{itemize}
    \item $\mathcal{K}$ is a set of closed formulae (i.e., axioms) defined by the variables, predicates and connectives in $\mathcal{L}$ and the set of domain symbols. They are highly interpretable propositional logic expressions.
    \item $\mathcal{G}(\cdot|\theta)$ is the parameter groundings of the symbols and logical operators,
    \item $ \Theta$ is the set of parameters in the groundings. This includes the trainable parameters of predicates and constants.
\end{itemize}

The training of an LTN model aims to find the set of optimal parameters $\Theta^*$ that maximise the aggregated satisfiability of $<\mathcal{K}, \mathcal{G}(\cdot|\theta)>$
\begin{align}
    \Theta^* &= \argmax_{\theta \in \Theta}\SatAgg_{\phi \in \mathcal{K}}(\mathcal{G}_{\theta}(\phi)) \\
    \SatAgg_{\phi \in \mathcal{K}}(\mathcal{G}_{\theta}(\phi)) &= 1-(\frac{1}{|\mathcal{K}|}\sum_{\phi \in \mathcal{K}}(\mathcal{G}_{\theta}(\phi)))^{\frac{1}{p}}
\end{align}

where $\text{SatAgg}(\cdot)$ is the function that aggregates the satisfiabilities of each axiom and $p$ is a hyperparameter. 

Therefore, the training goal can be formulated as the minimisation of the loss $\mathcal{L}$ :
\begin{equation}
    \mathcal{L} = 1-\SatAgg_{\phi \in \mathcal{K}}(\mathcal{G}_{\theta}(\phi))
\end{equation}

A more detailed and illustrative description of how these components are used to describe the logical relationship in medical ontologies is at \ref{pretraining}

 
\subsection{The Indication Relationship between Medications and Diagnoses}

If a medication $m$ was designed for treating a medical diagnosis $d$, an indication relationship $<m, d>$ can be defined. A medication can be designed to treat a set of medical conditions. If a medication is able to treat a parent node on the diagnosis ontology, it can be considered that it can cure all its children nodes. It is worth noticing that the indication relations graph does not enumerate all the existing indication relations.

\section{The OntoMedRec Model}

\subsection{Pre-training Ontology Encoders} \label{pretraining}
By definition, the chosen medical ontologies have the following characteristics:
\begin{itemize}
    \item \textbf{Explicit directed edges.} An edge in a medical ontology refers to a parent-child relationship between the two nodes. This relationship is not interchangeable or reflexive. 
    \item \textbf{Implicit deductive relationships.} Besides explicit edges in $\mathcal{E}_*$, there are deductive relationships in medical ontologies. 
    \begin{itemize}
        \item Two nodes with the same parent node are sibling nodes. They have definitive differences on their level.
        \item Ancestor nodes are multi-hop parent nodes. Ancestral relationships are not commutative or reflexive. We define one-hop ancestors as ``parents'' but not ``ancestors''. 
    \end{itemize} 
    \item Each node (except the root node) has only one parent. 
\end{itemize}

To accurately model the structural characteristics of medical ontologies, we pre-train three medical ontology node encoders, respectively for diagnoses, procedures and medications using logic tensor networks. Following the axioms used for the ontology deduction task in \cite{ltncap}, we devise a set of additional axioms regarding the explicit and deductive relationships among nodes. Additionally, we devise axioms to define the sibling relationships in the ontology. 


Since medical ontologies are much larger compared to the ontology in \cite{ltncap}, it is impractical to define variables over all the nodes in these three ontologies. For instance, to describe the axiom "the parent node of the parent node of a node is an ancestor node", three variables are required (``$\forall x,y,z: P(x,y) \land P(y,z) \to A(x,z)$'' where $x,y,z$ are variables and $P(\cdot,\cdot)$ and $A(\cdot,\cdot)$ are the predicates that calculate the satisfiability of the parent and ancestor relation). Each time a new variable is created over the entire ontology, a new copy of the embedding matrix of all the nodes ($\mathbf{E}_* \in \mathbb{R}^{|\mathcal{N}| \times d}$) is required, which is a task of the space complexity of $O(n|\mathcal{N}|d)$, where $n$ is the number of variables. To achieve efficient and effective training of the encoders, we design an axiom-oriented sampling method. We, firstly, randomly sample a batch of nodes from the ontology. Then, we sample all their respective ancestors, parents and siblings. This set of nodes constitutes a training node batch. All the directed edges between two nodes in the set constitute the positive edge samples, whereas all the node pairs without directive edges between them constitute the negative edge samples. With the adoption of the sampling method, the space complexity of the creation of variables are reduced to $O(nbd)$, where $b$ is the batch size.

\subsubsection{The Knowledge Formulation of Ontology Data}
Therefore, the knowledge of an ontology can be formulated as follows, using the notations  in \cite{ltn}. 
\begin{itemize}
    
\item \textbf{Domain} Medical terms in the ontology

\item \textbf{Variables}
$x$, $y$ and $z$, ranging over a batch of sampled nodes $N_b \subset \mathcal{T}_*$

\item \textbf{Predicates}
$P_{*}(x, y)$ as the parent scorer, $S_*(x,y)$ as the sibling scorer and $A_*(x,y)$ as the ancestor scorer

\item \textbf{Axioms}
\begin{itemize}
    \item Parental relationships are not reflexive and commutative: $\forall x \in N_{b}: \neg P_*(x, x)$, $\forall x, y \in N_{b}: P_*(x, y) \to \neg P_*(y, x)$
    \item Ancestral relationships are not reflexive and commutative: $\forall x \in N_{b}: \neg A_*(x, x)$, $\forall x, y \in N_{b}: A_*(x, y) \to \neg A_*(y, x)$
    \item The definition of sibling relationships (nodes with the same parent node): $\forall x,y,z \in N_{b}: P_*(x,y) \land P_*(x,z) \to S_*(y, z)$
    \item Sibling relationships are not reflexive but commutative: $\forall x \in N_{b}: \neg S_*(x, x)$, $\forall x, y \in N_{b}: S_*(x, y) \to S_*(y, x)$
    \item The parent node of a parent node is an ancestor node: $\forall x, y, z \in N_{b}: P_*(x,y) \land P_*(y,z) \to A_*(x, z)$
    \item The parent node of an ancestor node is an ancestor node: $\forall x, y, z \in N_{b}: P_*(x, y) \land A_*(y, z) \to A_*(x, z)$
    \item Positive and negative edges in the batch: $\forall (x, y) \in P_{b}: P_*(x, y)$, $\forall (x, y) \notin P_{b}: \neg P_*(x, y)$
\end{itemize}

\item \textbf{Grounding}
\begin{itemize}
    \item Let $\mathbf{v}_n$ be the representation of node $n$, $\mathcal{G}(\mathbf{v}_n)=\mathbb{R}^{d}$
    \item $\mathcal{G}(x|\theta)=\mathcal{G}(y|\theta)=\mathcal{G}(z|\theta)=[\mathbf{v}_n| n \in N_n]$
    \item $\mathcal{G}(P_*|\theta)$, $\mathcal{G}(S_*|\theta)$ and $\mathcal{G}(A_*|\theta)$ are $\sigma(\text{MLP}(x, y))$ with one output neuron and sigmoid function ($\sigma (\cdot)$) as the activation of the final layer
\end{itemize}
\end{itemize}

Ontology encoders are trained to maximise the aggregated satisfiability of all these axioms describing the structural characteristics of the ontology. For each ontology, we use a different set of predicates with the same structure. The three sets of predicates are optimised separately.


\subsubsection{The Alignment of Diagnosis and Medication Representations}

Intuitively, aligning the representations of medications and diagnoses after they have been respectively trained shortens the distance of these representations. The representations of diagnoses and medications are infused with the indication relationship. Therefore, using the pretrained representations from OntoMedRec as a starting point can improve the performance of the model, particularly in admissions with few-shot medications. Similarly, the knowledge of the MEDI dataset can be formulated as follows:



\begin{itemize}

\item \textbf{Domains: }Medical terms in the medication and diagnoses ontology

\item \textbf{Variables}
\begin{itemize}
    \item Medication $m$ ranging over all the medications in the batch of sampled indication pairs
    \item Diagnoses $s_x$ and $s_y$ ranging over all the medications in the batch of sampled indication pairs
\end{itemize}

\item \textbf{Predicates}
$I(m, d)$ for the indication relationship
\item \textbf{Axioms: } Let $\mathcal{I}$ be all the indication pairs in a sampled batch in MEDI dataset: $\forall (m, s_x) \in \mathcal{I}_b: I(m,s_x)$

\item \textbf{Grounding}
\begin{itemize}
    \item Let $\mathbf{m}$ and $\mathbf{d}$ and be the representation of medication $m$ and diagnosis $d$, $\mathcal{G}(\mathbf{m}) = \mathcal{G}(\mathbf{d}) =\mathbb{R}^{d}$
    \item $\mathcal{G}(m|\theta)=[\mathbf{m}_m| m \in \mathcal{I}]$, $\mathcal{G}(d|\theta)=[\mathbf{d}_d| d \in \mathcal{I}]$
    \item $\mathcal{G}(I|\theta)$ is $\sigma(\text{MLP}(m, d))$ with one output neuron and sigmoid function ($\sigma (\cdot)$) as the activation of the final layer
\end{itemize}
\end{itemize}

In each pretraining epoch, we train the three encoders sequentially then align medication and diagnosis embeddings with the indication dataset. We save the procedure embeddings with the highest satisfiability on the procedure ontology, and the medication and diagnoses embeddings with the highest satisfiability on the indication dataset.

\subsection{Fine-Tuning with Downstream Models}

Following the pre-training phase, the embeddings of medical terms are integrated with downstream medication recommendation models for further fine-tuning. We choose both instance-based models (Leap \cite{leap} and 4SDrug \cite{4sdrug}) and longitudinal models (RETAIN \cite{retain}, SafeDrug \cite{safedrug} and MICRON \cite{micron}) to fine-tune and evaluate OntoMedRec.

The representations of diagnoses and procedures (and medications, where possible) are loaded as a starting point for the respective embedding table and are further end-to-end fine-tuned with the medication recommendation task. 


\section{Experiments} \label{experiment}

\subsection{Experimental Setup}

\subsubsection{Dataset}

We use the ATC ontology for medications and the ICD9-CM ontology for diagnoses and procedures from BioPortal \cite{bioportal} to pretrain OntoMedRec. ICD9-CM is split into two sub-ontologies, respectively for diagnoses and procedures. The characteristics of these ontologies are listed in Table \ref{tab:ontologies}.

\begin{table}[h]
    \centering
    \caption{The statistical characteristics of the medical ontologies}
    \begin{tabular}{ l  c c c}
        \toprule
         & Diagnosis & Procedure & Medication\\
        \midrule
        \# nodes & 17737 & 4670 & 6441\\ 
         \midrule
         \# edges & 17736 & 4669 & 6440 \\ 
         \midrule
         Max depth & 7 & 4 & 5\\
         \bottomrule
    \end{tabular}
    \label{tab:ontologies}
\end{table}

\begin{table}[h]
    \centering
    \caption{The statistical characteristics of the MIMIC-III dataset}
    \begin{tabular}{l c} 
        \toprule
         Item & MIMIC-III \\
        \midrule
        \# patients & 35441\\ 
         \midrule
         \# admissions & 44129\\ 
         \midrule
         \# medications & 120\\
         \midrule
         \# procedures & 1975\\
         \midrule
         \# diagnoses & 6658\\
         \bottomrule
    \end{tabular}
    
    \label{tab:mimic-iii}
\end{table}

We use the benchmark dataset MIMIC-III \cite{mimic-iii} to fine-tune and evaluate the performance of downstream models integrated with the representations of OntoMedRec and other baselines. The statistical characteristics of the datasets are described in Table \ref{tab:mimic-iii}. To explore the performance of downstream models with or without OntoMedRec in sparse cases, we reserve the patients with only one admission, low-frequency diagnoses and low-frequency medications that were discarded in previous studies (e.g., in \cite{safedrug}). The ratio of training, testing and validation set is $4:1:1$. We split out a set of admissions with few-shot medications. We use TWOSIDES dataset \cite{twosides} as the ground truth of drug-drug interactions (DDIs). In contrast to previous studies, we reserve the drug pairs with lower numbers of DDIs.


\subsubsection{The Generation of Few-Shot Medications Test Cases}
We sort all medications in the EHR dataset by their frequencies. The medications with the lowest 30\% frequency (i.e., tail percentage) are designated as few-shot medications. Prescriptions in the test set with more than 1 few-shot medication are added to the few-shot test set.

\subsubsection{Baselines}
There are two major categories of existing medical ontology modelling methods: EHR-independent models (KAMPNet \cite{kampnet}) and EHR-dependent models (G-BERT \cite{gbert}). Both KAMPNet and G-BERT use GAT \cite{gat} to model medical ontology. Thus, we also choose GAT as one of the baselines to validate the effectiveness of our model. KAME \cite{kame} and GRAM \cite{gram} are not comparable to our models because their ontology training is deeply coupled with their downstream task which are not medication recommendation. GCN is commonly used for the modelling of EHR and DDI graphs \cite{gamenet}. Therefore, we choose randomly initiated naive embedding table, GAT \cite{gat} and GCN as baselines. GAT and GCN are fine-tuned along with the downstream models. We use link prediction as the pretraining task for these two models. The two baselines are trained 20 epochs, the best checkpoints with the lowest loss are selected. 

\subsubsection{Evaluation Metrics}
Following the evaluation protocol of many medication recommendation models, we use the following metrics:
\begin{itemize}
    \item \textbf{Jaccard coefficient. } It is the most common benchmark score for medication recommendation models. The Jaccard coefficient of all the the $n$ patient's $T$ admissions is calculated as follows:
    
    \begin{align} \label{jaccard}
    \text{Jaccard}^{(n)}_t &= \frac{|\{i: \mathbf{m}^{(n)}_{t,i}=1\} \cap \{i: \mathbf{\hat{m}}^{(n)}_{t,i}=1\}|}{|\{i: \mathbf{m}^{(n)}_{t,i}=1\} \cup \{i: \mathbf{\hat{m}}^{(n)}_{t,i}=1\}|} \\
    \text{Jaccard}^{(n)} &= \frac{1}{T^{(n)}}\sum_{t=1}^{T^{(n)}}\text{Jaccard}^{(n)}_t
    \end{align}
    where $\{i: \mathbf{m}^{(n)}_{t,i}=1\}$ is the set of indices where the element at $i$ on multi-hot encoding vector is 1. The higher the Jaccard coefficient is, the more accurate the recommendation is (i.e., the recommended set is more similar to the label).
    \item \textbf{Drug-Drug Interaction (DDI) score. } It is the percentage of medication pairs with known DDIs in the recommended set of medications. The lower it is, the fewer DDIs there are in the generated medication recommendation, and the safer the recommended medication combination can be considered. The DDI score of the $n$ patient's admissions is calculated as follows:
    \begin{equation}
        \text{DDI}^{(n)} = \frac{\sum_t^{T^{(n)}}\sum_{j,k \in \mathbf{\hat{m}}^{(n)}_{t,i}=1} \mathbf{1}\{\mathbf{D}_{j,k} =1\}}{\sum_{j,k \in \mathbf{\hat{m}}^{(n)}_{t,i}=1} 1}
    \end{equation}
    where $\mathbf{D} \in \mathbb{R}^{|\mathcal{M}| \times |\mathcal{M}|}$ is the DDI matrix retrieve from the TWOSIDES dataset \cite{twosides} and $\mathbf{1}\{\cdot\}$ is an indicator function that returns 1 when the input is true and 0 otherwise. $\mathbf{D}_{j,k} = 1$ indicates that medication $j$ and medication $k$ have at least one adverse effect when prescribed together.
    \item \textbf{F1 score. } It is commonly used as a metric for classification tasks. It is the harmonic mean of the precision and recall score and is calculated as follows:

    \begin{align}
     \text{Precision}^{(n)}_t &= \frac{|\{i: \mathbf{m}^{(n)}_{t,i}=1\} \cap \{i: \mathbf{\hat{m}}^{(n)}_{t,i}=1\}|}{|\{i: \mathbf{\hat{m}}^{(n)}_{t,i}=1\}|} \\
     \text{Recall}^{(n)}_t &= \frac{|\{i: \mathbf{m}^{(n)}_{t,i}=1\} \cap \{i: \mathbf{\hat{m}}^{(n)}_{t,i}=1\}|}{|\{i: \mathbf{m}^{(n)}_{t,i}=1\}|} \\
     \text{F1}^{(n)}_t &= 2 \cdot \frac{\text{Precision}^{(n)}_t \cdot \text{Recall}^{(n)}_t}{\text{Precision}^{(n)}_t + \text{Recall}^{(n)}_t} 
     \end{align}
\end{itemize}

\subsection{Results Discussion}
Table \ref{all_results} lists the results of the performance of downstream models in the entire testing set, and Table \ref{few_shots_results} lists the results with the few-shot medications testing set. The drop in performance between Table \ref{all_results} and Table \ref{few_shots_results} in all downstream models proves our assumption that the data sparsity issue harms the performance of downstream models. Although OntoMedRec cannot achieve the lowest DDI in some models in both test settings, they are lower than the ground truth DDI score (0.078 among the entire test set and 0.069 in the few-shot test set).

\begin{sidewaystable}[!htbp]
\centering
\captionsetup{justification=centering}
\caption{Performance of selected models on the MIMIC-III dataset.}
\begin{tabular*}{0.7\textheight}{l l c c c c c}
\toprule
\multicolumn{2}{c}{} & Jaccard & F1 & DDI & No. of drugs\\
\midrule
\multirow{4}{*}{LEAP} & Naive & $0.4689 \pm 0.0019$ & $0.6287 \pm 0.0019$ & $0.0603 \pm 0.0004$ & $17.8810 \pm 0.0405$ \\
& +GAT & $0.4178 \pm 0.0009$ & $0.5815 \pm 0.0009$ & $0.0631 \pm 0.0000$ & $19.9971 \pm 0.0014$ \\
& +GCN & $0.3853 \pm 0.0012$ & $0.5500 \pm 0.0013$ & $0.0769 \pm 0.0000$ & $12.9998 \pm 0.0002$ \\
& +OMR & $\mathbf{0.4732 \pm 0.0017}$ & $\mathbf{0.6322 \pm 0.0016}$ & $\mathbf{0.0596 \pm 0.0004}$ & $17.1709 \pm 0.0465$ \\
\midrule
\multirow{4}{*}{SafeDrug} & Naive & $0.5431 \pm 0.0016$ & $0.6920 \pm 0.0014$ & $0.0600 \pm 0.0002$ & $21.9985 \pm 0.0668$ \\
& +GAT & $0.5232 \pm 0.0022$ & $0.6740 \pm 0.0020$ & $\mathbf{0.0569 \pm 0.0002}$ & $21.6908 \pm 0.0838$ \\
& +GCN & $0.5202 \pm 0.0013$ & $0.6707 \pm 0.0010$ & $0.0580 \pm 0.0002$ & $22.1046 \pm 0.1074$ \\
& +OMR & $\mathbf{0.5481 \pm 0.0024}$ & $\mathbf{0.6965 \pm 0.0021}$ & $0.0589 \pm 0.0002$ & $22.1837 \pm 0.0790$ \\
\midrule
\multirow{4}{*}{MICRON} & Naive & $0.5147 \pm 0.0020$ & $0.6645 \pm 0.0020$ & $\mathbf{0.0504 \pm 0.0005}$ & $15.7760 \pm 0.1218$ \\
& +GAT & $0.4991 \pm 0.0026$ & $0.6508 \pm 0.0023$ & $0.0510 \pm 0.0003$ & $15.4560 \pm 0.1239$ \\
& +GCN & $0.5003 \pm 0.0017$ & $0.6524 \pm 0.0016$ & $0.0511 \pm 0.0004$ & $15.4916 \pm 0.0648$ \\
& +OMR & $\mathbf{0.5203 \pm 0.0019}$ & $\mathbf{0.6696 \pm 0.0019}$ & $0.0517 \pm 0.0003$ & $16.1107 \pm 0.1218$ \\
\midrule
\multirow{4}{*}{4SDrug} & Naive & $0.4667 \pm 0.0017$ & $0.6261 \pm 0.0016$ & $0.0478 \pm 0.0004$ & $13.7054 \pm 0.0463$ \\
& +GAT & $0.4666 \pm 0.0016$ & $0.6260 \pm 0.0017$ & $0.0477 \pm 0.0005$ & $13.7442 \pm 0.0624$ \\
& +GCN & $\mathbf{0.4670 \pm 0.0017}$ & $\mathbf{0.6263 \pm 0.0016}$ & $0.0478 \pm 0.0004$ & $13.7361 \pm 0.0559$ \\
& +OMR & $0.4662 \pm 0.0015$ & $0.6257 \pm 0.0014$ & $\mathbf{0.0474 \pm 0.0005}$ & $13.7321 \pm 0.0724$ \\
\midrule
\multirow{4}{*}{RETAIN} & Naive & $0.5433 \pm 0.0023$ & $0.6913 \pm 0.0019$ & $0.0646 \pm 0.0006$ & $17.0477 \pm 0.1014$ \\
& +GAT & $0.4264 \pm 0.0023$ & $0.5871 \pm 0.0023$ & $\mathbf{0.0554 \pm 0.0006}$ & $14.9248 \pm 0.0877$ \\
& +GCN & $0.4335 \pm 0.0020$ & $0.5909 \pm 0.0018$ & $0.0574 \pm 0.0006$ & $16.4566 \pm 0.0917$ \\
& +OMR & $\mathbf{0.5536 \pm 0.0019}$ & $\mathbf{0.7001 \pm 0.0015}$ & $0.0642 \pm 0.0005$ & $17.7567 \pm 0.0836$ \\
\bottomrule
\end{tabular*}
\label{all_results}

\end{sidewaystable}

\begin{sidewaystable}[!htbp]
\centering
\caption{Performance of selected models on the test set with few-shot medications.}
\begin{tabular*}{0.7\textheight}{l l c c c c c}
\toprule
\multicolumn{2}{c}{} & Jaccard & F1 & DDI & No. of drugs\\
\midrule
\multirow{4}{*}{Leap} & Naive & $0.4328 \pm 0.0060$ & $0.5981 \pm 0.0060$ & $0.0522 \pm 0.0018$ & $18.3266 \pm 0.1629$ \\
& +GAT & $0.3978 \pm 0.0064$ & $0.5636 \pm 0.0067$ & $0.0632 \pm 0.0000$ & $20.0000 \pm 0.0000$ \\
& +GCN & $0.3451 \pm 0.0046$ & $0.5083 \pm 0.0051$ & $0.0769 \pm 0.0000$ & $13.0000 \pm 0.0000$ \\
& +OMR & $\mathbf{0.4341 \pm 0.0071}$ & $\mathbf{0.5986 \pm 0.0071}$ & $\mathbf{0.0571 \pm 0.0023}$ & $17.4800 \pm 0.2528$ \\
\midrule
\multirow{4}{*}{SafeDrug} & Naive & $0.5141 \pm 0.0083$ & $0.6705 \pm 0.0074$ & $0.0577 \pm 0.0007$ & $23.8330 \pm 0.5112$ \\
& +GAT & $0.5044 \pm 0.0054$ & $0.6618 \pm 0.0049$ & $0.0564 \pm 0.0010$ & $23.8990 \pm 0.2941$ \\
& +GCN & $0.4940 \pm 0.0082$ & $0.6517 \pm 0.0079$ & $\mathbf{0.0562 \pm 0.0009}$ & $24.6039 \pm 0.4459$ \\
& +OMR & $\mathbf{0.5206 \pm 0.0071}$ & $\mathbf{0.6769 \pm 0.0062}$ & $0.0587 \pm 0.0011$ & $24.7879 \pm 0.4228$ \\
\midrule
\multirow{4}{*}{MICRON} & Naive & $0.4849 \pm 0.0099$ & $0.6411 \pm 0.0101$ & $0.0615 \pm 0.0018$ & $20.5828 \pm 0.6436$ \\
& +GAT & $0.4626 \pm 0.0061$ & $0.6215 \pm 0.0067$ & $\mathbf{0.0583 \pm 0.0023}$ & $19.9857 \pm 0.9167$ \\
& +GCN & $0.4660 \pm 0.0062$ & $0.6240 \pm 0.0052$ & $0.0628 \pm 0.0010$ & $20.0187 \pm 0.6505$ \\
& +OMR & $\mathbf{0.4876 \pm 0.0094}$ & $\mathbf{0.6428 \pm 0.0087}$ & $0.0622 \pm 0.0012$ & $20.6576 \pm 0.5616$ \\
\midrule
\multirow{4}{*}{4SDrug} & Naive & $0.4310 \pm 0.0071$ & $0.5953 \pm 0.0064$ & $0.0385 \pm 0.0020$ & $14.4564 \pm 0.2999$ \\
& +GAT & $0.4298 \pm 0.0064$ & $0.5947 \pm 0.0062$ & $0.0385 \pm 0.0016$ & $14.3754 \pm 0.3247$ \\
& +GCN & $0.4304 \pm 0.0068$ & $0.5949 \pm 0.0065$ & $0.0380 \pm 0.0023$ & $14.2446 \pm 0.3319$ \\
& +OMR & $\mathbf{0.4316 \pm 0.0079}$ & $\mathbf{0.5961 \pm 0.0075}$ & $\mathbf{0.0379 \pm 0.0018}$ & $14.4943 \pm 0.3259$ \\
\midrule
\multirow{4}{*}{Retain} & Naive & $0.5057 \pm 0.0077$ & $0.6615 \pm 0.0069$ & $0.0608 \pm 0.0018$ & $19.7118 \pm 0.6843$ \\
& +GAT & $0.3650 \pm 0.0076$ & $0.5283 \pm 0.0081$ & $\mathbf{0.0585 \pm 0.0014}$ & $16.1552 \pm 0.2670$ \\
& +GCN & $0.3637 \pm 0.0069$ & $0.5263 \pm 0.0076$ & $0.0700 \pm 0.0011$ & $18.2791 \pm 0.3432$ \\
& +OMR & $\mathbf{0.5229 \pm 0.0089}$ & $\mathbf{0.6780 \pm 0.0078}$ & $0.0609 \pm 0.0015$ & $20.6692 \pm 0.6741$ \\
         
\bottomrule
\end{tabular*}
    
\label{few_shots_results}
\end{sidewaystable}

\newpage

\subsubsection{Results on the entire MIMIC-III dataset}

As we can observe from Table \ref{all_results}, integrating the representations of diagnoses, procedures and medications (where possible) from OntoMedRec can improve the performance of most selected medication recommendation models in the entire dataset compared to all baselines. This demonstrates the representation of OntoMedRec is model-agnostic for downstream medication recommendation models. 

\subsubsection{Results on the few-shot cases}

The representations of OntoMedRec improve all compared downstream models in the few-shot medication test set. We can observe from the result that the representations improve the performance for all compared longitudinal models. For compared instance-based models, although the representations of OntoMedRec do not improve their performance by a large margin, we notice that 1) they have lower performance scores compared to selected longitudinal models, and 2) the performance after the integration of OntoMedRec is not lower than the best performance by a large margin. We speculate the reason is that instance-based models adopted fewer pretrained embeddings comparing to longitudinal models.


\subsubsection{Further Investigation of Sparse Scenarios}
\begin{figure}[H]
    \centering
    \includegraphics[width=0.7\linewidth]{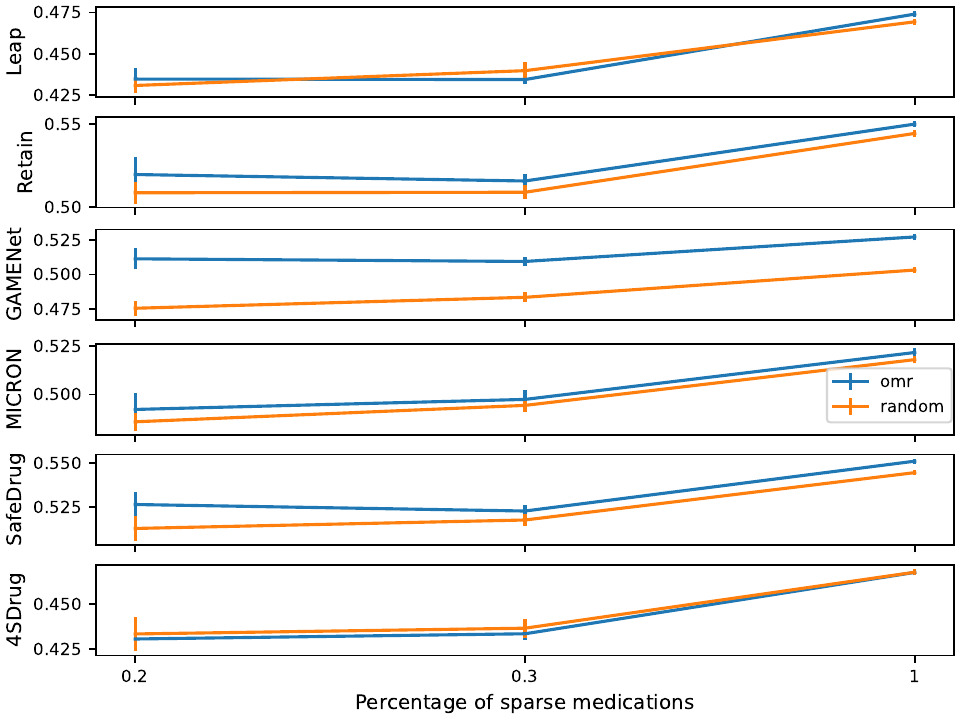}
    \caption{Performance of randomly initialised embedding table and OntoMedRec embedding table in different downstream models and tail percentages}
    \label{fig:result_vis}
\end{figure}

To further investigate how few-shot medications affect the performance of medication recommendation models with OntoMedRec, we further compare OntoMedRec representations and randomly initialised embedding table with different sparse settings starting with the lowest frequency being 20\% (as visualised in Fig.\ref{fig:result_vis}). The representation of OntoMedRec can improve the performance of longitudinal downstream models in all three test sets. Overall, the performance gap between models with OntoMedRec pretraining and models without pretraining is larger when the data is sparser (20\% is the sparsest scenario) which shows that medication recommendation models can benefit more from OntoMedRec in sparser scenarios. For LEAP, OntoMedRec can improve its performance for the entire testing set and the test set with the 20\%-least-frequent medications. For 4SDrug, the performance margin is small. 

\section{Conclusion}

In this paper, we proposed OntoMedRec, the self-supervised, logically-pretrained model-agnostic ontology encoders for medication recommendation. We devise axioms that collectively define the structure of medical ontologies, and use logical tensor networks (LTNs) to maximise the satisfiability of the representations. Furthermore, we align the representations of diagnoses and medications with medication indication information. The ontology-enhanced representation can be integrated into various downstream medication recommendation models to alleviate the negative effect brought by the data sparsity issue. We conducted experiments to evaluate the efficacy of OntoMedRec. Results show that the representation of OntoMedRec can improve the performance of most selected models in the entire testing dataset, and that it can improve the performance of all longitudinal models in the few-shot medications test set.


%
%
%

\bibliography{sn-bibliography}

\section{Declarations}
\subsection{Ethical Approval}
not applicable

\subsection{Funding}
This paper is funded by the Graduate Research Industry Partnership (GRIP) program. More information can be found here: https://www.monash.edu/msdi/study/graduate-research-program/engagement-opportunities/graduate-research-industry-partnership-grip-program. 

\subsection{Availability of Data and Materials}
We use the ATC ontology for medications and the ICD9-CM ontology for diagnoses
and procedures from BioPortal \cite{bioportal} to pretrain OntoMedRec. We use TWOSIDES dataset \cite{twosides} as the ground
truth of drug-drug interactions (DDIs). These three datasets are all publicly available. 
We use the benchmark dataset MIMIC-III \cite{mimic-iii} to fine-tune and evaluate the perfor-
mance of downstream models integrated with the representations of OntoMedRec and
other baselines. It is a free-to-use large medical records dataset that can be accessed here: https://physionet.org/content/mimiciii/1.4/. It can be accessed by credentialed users upon the completion of online ethics training and the signing of the user agreement. 




\end{document}